\pdfoutput=1

\documentclass[11pt]{article}

\usepackage[final]{acl}

\usepackage{times}
\usepackage{latexsym}

\usepackage[T1]{fontenc}

\usepackage[utf8]{inputenc}

\usepackage{microtype}

\usepackage{inconsolata}
\usepackage{booktabs}

\usepackage{graphicx}
\usepackage{multirow}
\usepackage{subcaption}
\usepackage{graphicx}
\usepackage{amsmath}
\usepackage{amssymb}
\usepackage{mathtools}
\usepackage{amsthm}
\usepackage{url}
\usepackage{hhline}
\usepackage{caption}
\usepackage{tcolorbox}
\title{Complete Chess Games Enable LLM Become A Chess Master}

\author{
 \textbf{Yinqi Zhang},
 \textbf{Xintian Han},
 \textbf{Haolong Li},
 \textbf{Kedi Chen},
 \textbf{Shaohui Lin\thanks{Corresponding Author}}
}

\begin{document}
\maketitle

\begin{abstract}
Large language models (LLM) have shown remarkable abilities in text generation, question answering, language translation, reasoning and many other tasks. It continues to advance rapidly and is becoming increasingly influential in various fields, from technology and business to education and entertainment. Despite LLM's success in multiple areas, its ability to play abstract games, such as chess, is underexplored. Chess-playing requires the language models to output legal and reasonable moves from textual inputs. Here, we propose the Large language model ChessLLM to play full chess games. We transform the game into a textual format with the best move represented in the Forsyth-Edwards Notation. We show that by simply supervised fine-tuning, our model has achieved a professional-level Elo rating of 1788 in matches against the standard Elo-rated Stockfish when permitted to sample 10 times. We further show that data quality is important. Long-round data supervision enjoys a 350 Elo rating improvement over short-round data.
\end{abstract}
\section {Introduction}
Recently, Large Language Models (LLMs) based on transformer architectures~\cite {vaswani2017attention} have demonstrated capabilities well beyond language modeling. A key milestone was the advent of ChatGPT~\cite {ouyang2022training}. Extensive research has focused on developing efficient LLM base models~\cite {du2021glm,biderman2023pythia,black2022gpt,together2023redpajama,touvron2023llama}, including supervised models~\cite {taori2023stanford,chiang2023vicuna,gpt4all,kopf2023openassistant} and models using Reinforcement Learning from Human Feedback (RLHF)~\cite {christiano2017deep,ouyang2022training,rando2023universal,bai2023qwen}. Recent research~\cite {wei2022emergent,li-etal-2024-exploring-mathematical} shows that as models scale, their capabilities increase. This raises questions about LLMs' intelligence and learning structures.
Chess, an ancient game, has dialogue-like characteristics in its notational structures such as Forsyth-Edwards Notation (FEN), Standard Algebraic Notation (SAN), and Universal Chess Interface (UCI). Machine learning in chess has evolved to include reinforcement learning and neural networks based on supervised learning from human gameplay. Developments include AI-based engines like Leela Chess Zero (LC0)\footnote{\href{https://lczero.org}{https://lczero.org}} and Stockfish NNUE\footnote{\href{https://stockfishchess.org}{https://stockfishchess.org}}, which refine their algorithms through new learning. Deep learning has shown the potential of AI in strategic games. The ChessGPT~\cite{feng2023chessgpt} model demonstrated the ability to choose optimal moves by learning from human language and chess data. However, models like ChessGPT cannot generate the best move based on the current game state and complete an entire match. Our focus is on match completeness and quality of gameplay.

Our contributions can be listed as follows:
\begin{itemize}
\item{\textbf{Dataset.} We collected a large dataset of chess games with over 20B tokens from open-source platforms. Data quality matters; long round data supervision outperforms short-round data by 350 Elo points.}
\item{\textbf{Model.} Our ChessLLM is designed to play entire chess games through dialogues. After fine-tuning, it achieved an Elo rating of 1788, winning 61\% of games against Stockfish at skill level 0, 56\% at skill level 1, and 30\% at skill level 2. }
\item{\textbf{Eval Method.} We propose evaluation methods based on full games against Stockfish, including move validity, Elo rating, and win rate.
We are the only ones using a large language model for chess that can complete full games.}
\end{itemize}

\begin{figure*}[htbp]
\centering
\vspace{-2.5em}
\includegraphics[width=\linewidth]{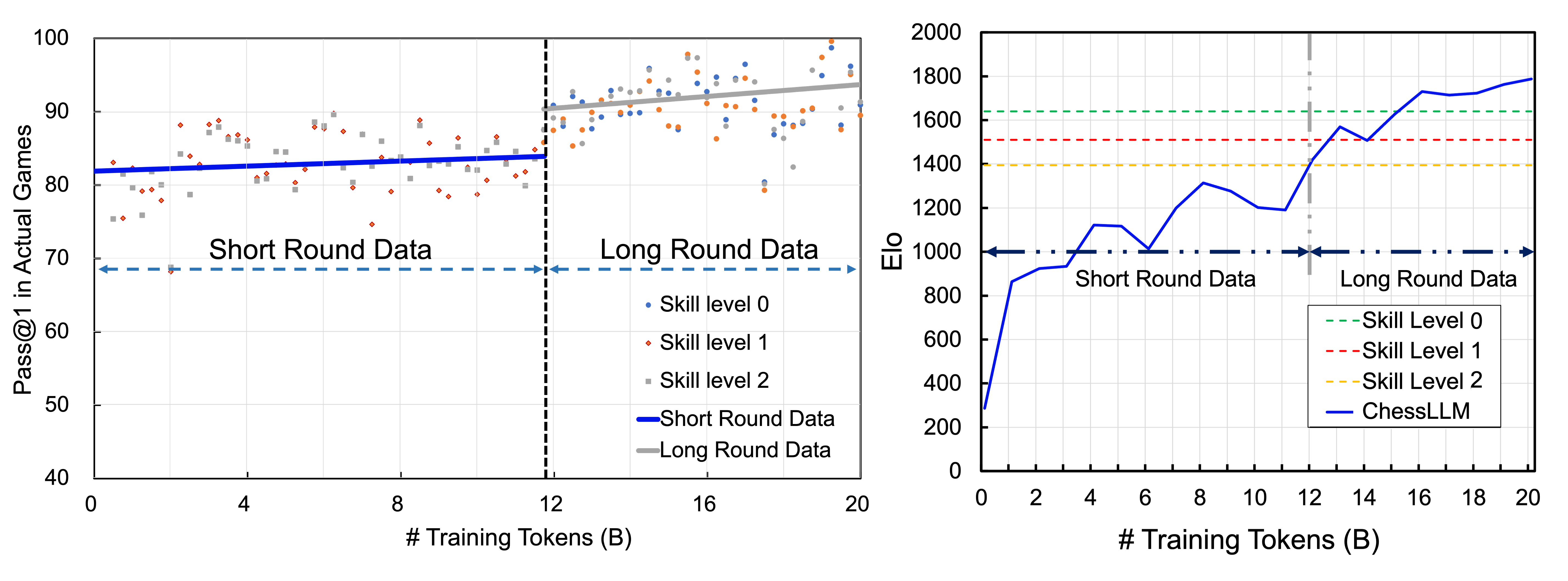}
\vspace{-2.5em}
\caption{\textbf{Left: $pass@1$} increases with the number of tokens. After introducing long-round data, $pass@1$ further increases. \textbf{Right: }The Elo Rating of ChessLLM with the number of training tokens. Skill level indicates the level of Stockfish.}
\label{fig:elototal}
\vspace{-1em}
\end{figure*}
\section{Related work}
\subsection{Large Language Model}
The emergence of Large language models (LLMs) GPT-4~\cite{achiam2023gpt}, stands as a noteworthy testament to the significant advancements in natural language understanding and generation.
Unlike commercial models, open-source models, such as Alpaca~\cite{alpaca}, Vicuna~\cite{zheng2023judging}, and Llama2~\cite{touvron2023llama2}, have recently become more accessible.
Due to their proficiency in text reasoning, LLMs are increasingly being utilized in everyday applications\citep{chen2024diahaludialoguelevelhallucinationevaluation}. Comprehensive benchmarks, such as MMLU~\cite{hendryckstest2021} and HELM~\cite{lee2023holistic}, have been devised for thorough assessments of the LLMs' overall capabilities. Our work takes this evaluation process one step further, particularly highlighting and investigating the capacity of LLMs' ability to play abstract games.

\subsection{Supervised Fine-tuning}

Supervised Fine-tuning has emerged as a revolutionary technique within the field of machine learning and has been the subject of a multitude of studies. Owing to the continuous advancements in the domain of transfer learning, pre-trained models, fine-tuned in a supervised manner, have demonstrated superior performance in numerous tasks. Notably, in the context of natural language processing (NLP), the work by Howard and Ruder became a pioneering model of this technique. Their method~\cite{howard2018universal} leverages the power of transfer learning for comprehensive language modeling tasks, thus effectively surpassing previous benchmarks.
Manipulating the same concept, BERT~\cite{devlin2019bert}, an innovative model fine-tuned in a supervised manner for a wide array of NLP tasks. BERT demonstrated remarkable success within various NLP tasks, setting new performance standards.

In this work, we trained ChessLLM with supervised fine-tuning.
\subsection{Chess}

The quest to develop artificial intelligence capable of playing chess can be traced back to the inception of computer science~\cite{turing1953digital,campbell2002deep}. The application of machine learning, particularly deep learning, in the domain of chess has been explored extensively in recent years~\cite{silver2018general,mcgrath2022acquisition}. One of the pivotal works in this field is the study by DeepChess~\cite{David_2016}, which presented an end-to-end learning method for chess based solely on deep neural networks, demonstrating the powerful capabilities of machine learning in comprehending and mastering strategic games without a priori knowledge.



In this work, we applied LLMs to chess and evaluated them with Elo rating.

\section {A Large Scale Dataset of Chess}\label {sec:dataset}
We introduce a large-scale dataset by collecting chess games online and generating the best moves based on Stockfish's evaluations.
Previous research relied on Portable Game Notation (PGN) for strategy learning, interpreting moves as actions in a Markov Decision Process. ChessGPT sees additional value in PGN data, such as Elo ratings indicating player strength and annotated moves providing computer-generated evaluations. These annotations aid in value function learning, thus ChessGPT retains all this information for easier strategy learning.
We argue that the core of chess is making the best decision for a given Forsyth-Edwards Notation (FEN) position. Human players focus on the current position rather than past moves. While ChessGPT uses historical moves, formats like PGN can be inefficient for large language models (LLMs) due to their expanding token length. The FEN format remains constant, making it more suitable for LLMs. Therefore, we constructed our dataset as FEN-Best move pairs.
\paragraph{Best Move Construction}Our Best Move dataset was created through a search method using Stockfish. It consists of two parts: the short round dataset from Chessdb\footnote{\href{http://chessdb.sourceforge.net}{http://chessdb.sourceforge.net}} and the long round dataset from self-play endgames based on Stockfish evaluations. Stockfish evaluates positions using heuristic functions and an alpha-beta game tree search. We searched for valid moves from current positions, with search depths of 12-50 for short rounds and 50-200 for long rounds, limiting each search to two seconds. The highest win-rate moves were selected as the best moves.

\section {Model}
\begin{figure}[htbp]
    \centering
    \includegraphics[width=0.4\textwidth]{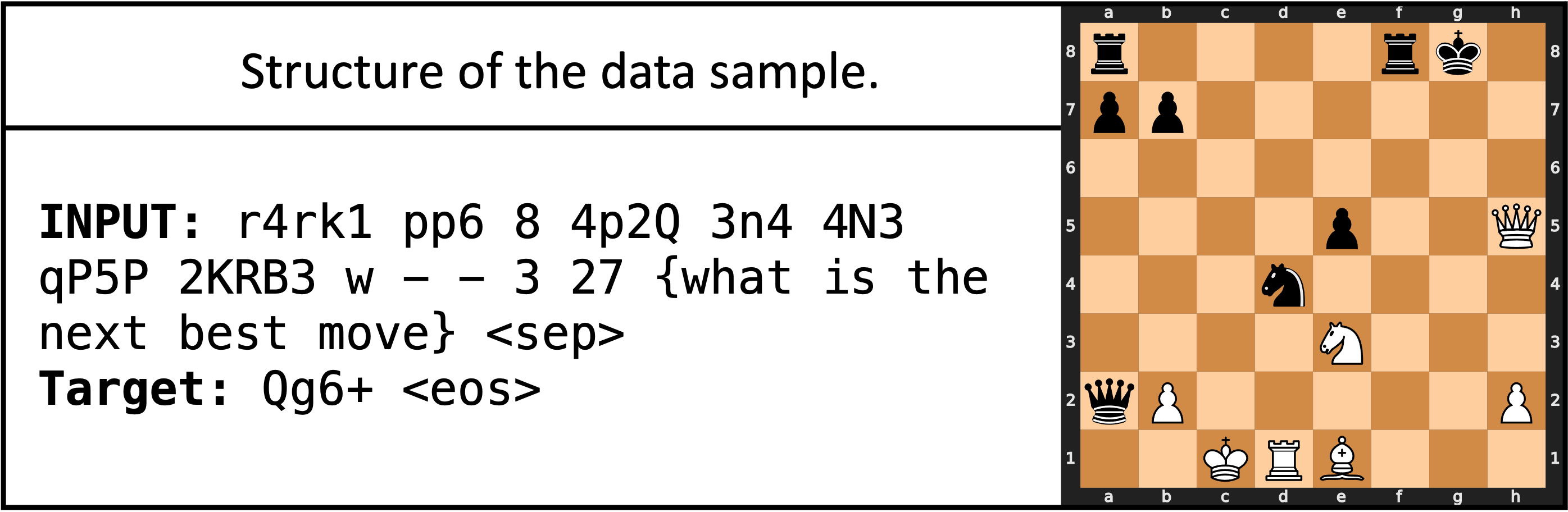}
    \caption{One example of training data.}
    \label{fig:datasample}
\end{figure}

The Generative Pre-trained Transformer (GPT-3) is an autoregressive language model that generates human-like text through deep learning. It trains on casual language modeling, predicting the next word based on previous words. We trained a GPT-like model using open-llama-3B~\cite {openlm2023openllama} and the chess resources from Section~\ref {sec:dataset}.
Unlike policy behavior data in robotics or gaming, chess state and move data can be expressed textually. This allows chess to be rendered as a text-based game, enabling imitation learning for policy through casual language modeling of the game dataset (Figure~\ref{fig:datasample}).
This innovative approach of applying language modeling to chess signifies a novel shift in policy learning, leveraging the game's unique aspects to develop superior gameplay tactics. 

\section {Evaluation Methods}
Chess requires a dynamic evaluation method beyond a fixed set typical of NLP tasks. We propose supplementing the evaluation set with actual games to better assess the model's capabilities.
\subsection {Actual Games}
Playing against Stockfish, a top chess engine, offers a strategic challenge. Stockfish uses advanced algorithms to determine optimal moves. Players can choose time controls (blitz, quick, or traditional) to set the gameplay tempo. The engine analyzes moves and positions to find the best move using its evaluative function.
In our experiments, we analyzed metrics such as $pass@1$ and win rate. We believe using Stockfish against our model more authentically simulates real-world human-model interactions and offers greater robustness than a static evaluation set.
\begin{figure*}[t]
    \centering
    \vspace{-2em}
    \includegraphics[width=\textwidth]{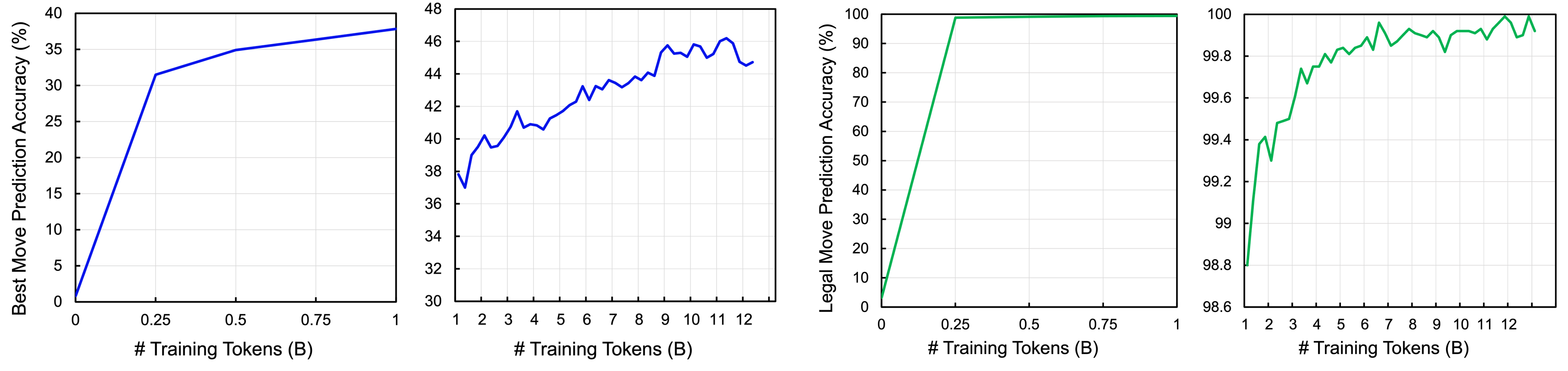}
    \vspace{-1.em}
    \caption{\textbf{Left: }Best Move Accuracy of ChessLLM training with short round data. The accuracy of the best move increases with the number of training tokens. \textbf{Right: }Legal Move Accuracy of ChessLLM training with short round data. The accuracy of the legal move increases with the number of training tokens. }
    \label{fig:midlegal}
\end{figure*}
\begin{figure}[t]
    \centering
    \vspace{-.8em}
    \includegraphics[width=\linewidth]{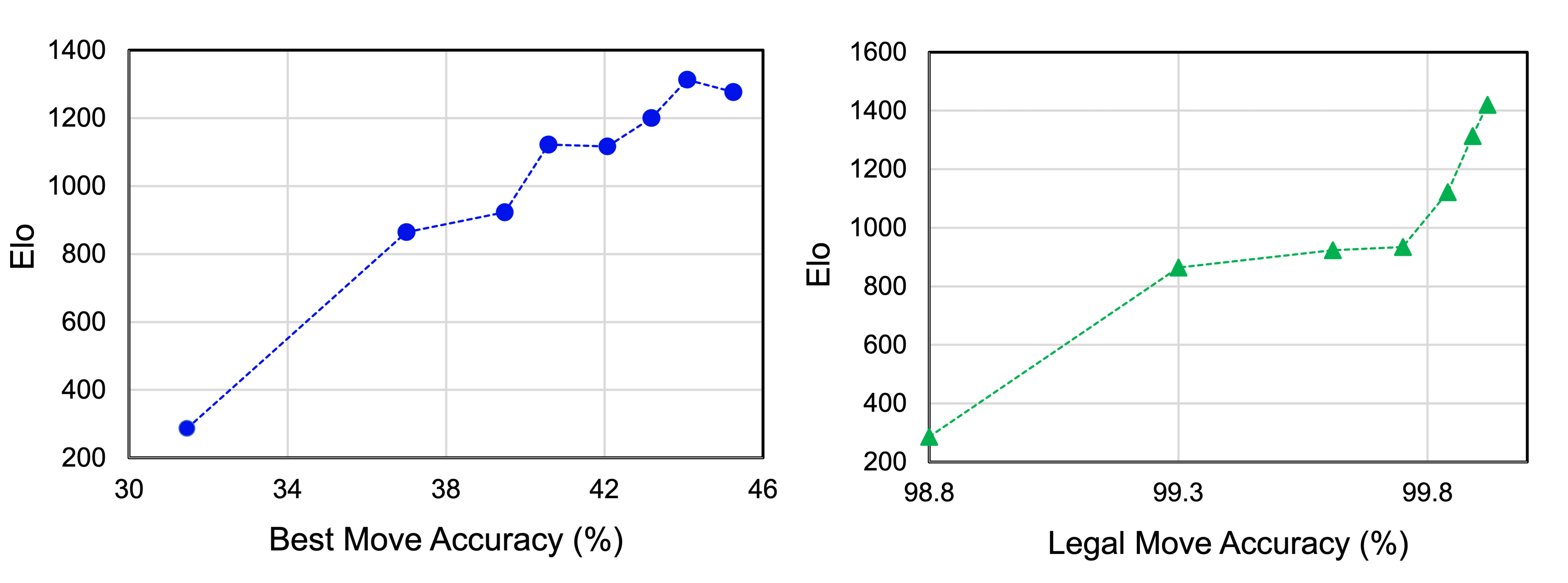}
    \vspace{-2em}
    \caption{\textbf{Left: }Correlation between ChessLLM's best move accuracy and its Elo rating. \textbf{Right: }Correlation between ChessLLM's legal move accuracy and its Elo rating.}
    \label{fig:elo-best}
    \vspace{-1.5em}
\end{figure}
\noindent
\paragraph{Pass@1 in Actual Games.}We evaluated our model's performance across different data scales, focusing on its ability to generate legal moves successfully.

\noindent
\paragraph{Win Rating.}The win rating refers to victories, draws, and losses out of 100 rounds when the model competes against Stockfish or other engines.

\noindent
\paragraph{Elo Rating.}We ran a series of matches between our model and Stockfish, recording strategies and moves. The Elo rating is calculated using the formula
\begin{equation}
    Elo_N = Elo_O + (R_A-R_E)K,\\
    \label{stockfish_elo}
\end{equation}
\begin{equation}
    R_E = \frac{1}{1 + 10^{\frac{Elo_{S} - Elo_{M}}{400}}}.
\end{equation}
where 
$Elo_N$ is the updated Elo rating after the game.
$Elo_O$ is the previous Elo rating before the game.
$K$ is the weight of the tournament. In professional chess, $K$ is often set to 10 for high-ranked players and 20 for low-ranked players.
$R_A$ is the actual result of the game (1 for win, 0.5 for draw, 0 for loss).
$R_E$ is the expected result of the game.
$Elo_{S}$ is the old Elo rating of Stockfish.
$Elo_{M}$ is the old Elo rating of the model. Moreover, we refer to the method introduced by Stockfish to convert between its skill level and Elo rating. The specific calculation method is shown as follows.
\begin{equation}
    SK = 37.247e^3 - 40.852e^2 + 22.294e - 0.311,
    \label{eq:sk}
\end{equation}
\begin{equation}
   e = \frac{Elo-1320}{1870},
   \label{eq:elos}
\end{equation}
where $SK$ represents skill level $SK = 0,1,2,...,20$, and $Elo$ represents Stockfish’s Elo rating.
\subsection{Evaluation Set}
While games against Stockfish provide a robust performance assessment, their length introduces substantial evaluation costs. Thus, we also use an evaluation set to measure the model's prowess.
Data distribution in the evaluation set focuses on games spanning 10-20 rounds (30\%) and 20-40 rounds (50\%), emphasizing the model's middle-game capabilities. This approach manages the inherent uncertainty in chess match lengths, ensuring the model does not exhibit forgetting phenomena after exposure to long rounds.
\noindent
\paragraph{Distribution of Training set and Evaluation set}Our training data was generated with $depth=1$ and $time limited=0.1$, while the data used in the game process was generated with $time limited=10$ and without depth limited. The eval set is produced by $depth=1$ and $time limited=0.1$, the same as the train set. These two datasets are from different domains, so our method is effective not only on in-domain data.
\noindent
\paragraph{Legal Move Accuracy.}We used Stockfish to generate legal move responses for 10,000 unique board positions not in the training set, evaluating our model's proposed moves for legality to ensure proper convergence.

\noindent
\paragraph{Best Move Accuracy.}Stockfish generated best move responses, allowing us to compare its outcomes with our model to calculate the accuracy rate for best move predictions.

\section{Experiment Analysis}


        


\subsection{Evaluation Set}
\begin{table}[t]
\caption{Exhibition of Match Results and Computed Elo Scores of ChessLLM \textit{vs.} Stockfish at Different Skill Levels. The table enumerates the number of wins, losses, and draws, along with the calculated Elo scores of ChessLLM when competing against Stockfish at varying skill levels.}
    \vspace{-.5em}
    \centering
     \resizebox{\linewidth}{!}{
    \begin{tabular}{c|cc|cccc}
    \toprule
        \multirow{2}{*}{} &  \multicolumn{2}{c|}{Stockfish} & \multicolumn{4}{c}{ChessLLM}\\
        & Skill level&Elo&Win& Lose &Draw & Elo \\\hline
       ChessLLM   & 0 & 1350-1440&61&29&10&1632 $\pm$ 45\\
           \textit{vs.}          & 1 & 1450-1560&56&37&7&1753 $\pm$ 55\\
        Stockfish        & 2 & 1570-1720&30&69&1&1788 $\pm$ 75\\    
    \bottomrule

    \end{tabular}
    }
        \vspace{-.5em}

    \label{tab:elo chessegpt}
\end{table}
\begin{table}[t]
    \centering
    \caption{General policy evaluation in Black. Note LLAMA denotes the LLAMA-7B}
        \vspace{-.5em}
            \resizebox{\linewidth}{!}{

    \begin{tabular}{c|ccccc}\toprule
        \multirow{2}{*}{Elo Rating} & \multicolumn{5}{c}{Move Scores (\%)} \\
         & LLAMA&RedPajama&ChessGPT-Base&ChessGPT-Chat&ChessLLM\\\hline
         700-1000 & 52.9 $\pm$ 0.9 & 46.2 $\pm$ 1.0 & 51.9 $\pm$ 0.1 & 52.1 $\pm$ 0.9 & \textbf{90.96 $\pm$ 1.4}\\
        1200-1500 & 53.2 $\pm$ 0.9 & 46.9 $\pm$ 0.9 & 53.0 $\pm$ 1.0 & 52.4 $\pm$ 1.0 & \textbf{95.11 $\pm$ 0.8}\\
        1700-2000 & 52.1 $\pm$ 0.8 & 46.6 $\pm$ 1.0 & 52.0 $\pm$ 1.0 & 52.0 $\pm$ 1.0 & \textbf{96.88 $\pm$ 0.9}\\
        2700-3000 & 53.6 $\pm$ 0.9 & 47.3 $\pm$ 1.0 & 52.2 $\pm$ 0.9 & 52.1 $\pm$ 1.1 & \textbf{97.14 $\pm$ 0.6}\\
        \bottomrule
        
    \end{tabular}}
    \vspace{-1em}

    \label{tab:gpcompare}
\end{table}
\begin{table}[t]
    \centering
        \caption{ The win rates of various LMs when competing in Chess. Note LLAMA denotes the LLAMA-7B.}
    \resizebox{\linewidth}{!}{
    \begin{tabular}{c|cccc}\toprule
            & LLAMA     &RedPajama&ChessGPT-Base&ChessGPT-Chat\\\hline
LLAMA           &-                &-             &-            &-\\
RedPajama           &22.2 $\pm$ 4.2&-             &-            &-\\
ChessGPT-Base  &61.3 $\pm$ 2.4&73.6 $\pm$ 1.1&-            &-\\
ChessGPT-Chat  &59.8 $\pm$ 1.5&70.8 $\pm$ 0.7&48.8 $\pm$ 2.7&-\\
ChessLLM(Ours)       &\textbf{89.8 $\pm$ 0.8}&\textbf{95.5 $\pm$ 0.1}&\textbf{91.7 $\pm$ 0.3}&\textbf{92.3 $\pm$ 0.1}\\\bottomrule
    \end{tabular}}
    \vspace{-1em}
    \label{tab:winrate}
\end{table}

We evaluated in-distribution data to analyze our model's performance on the evaluation set under varying computing power. From Fig.~\ref {fig:midlegal}, we observed that on in-distribution data, model performance improves with an increase in training tokens, but at a diminishing rate. This relationship is crucial for understanding model scalability and resource allocation during training. Note that "same distribution" refers to the FEN board state distribution and its corresponding best move.


\noindent
\paragraph{Legal Move and Best Move Accuracy.}Fig.~\ref{fig:midlegal} Left shows that with only 0.5B tokens, our model achieves a legal move accuracy of 99.11\% on in-distribution boards, indicating its impressive preliminary chess playing ability. As data volume increases, performance improves, demonstrating the model's scalability and potential for further enhancement. The high accuracy with just 0.5B tokens underscores the model's efficiency and effectiveness. Fig.~\ref{fig:midlegal} Right shows the Best Move accuracy under the same distribution. With 2.75B tokens, the model achieved a Best Move accuracy of 40.11\%. Although the logic is similar, the generation steps differ, highlighting our model's ability to accurately predict the best moves in most cases, proving its practical utility.


\subsection{Actual Games}

\paragraph{Pass@1 in Actual Games.}The $temperature$ and $top_p$ parameters were both set at $1.0$, and $top_k$ was set at $50$ we generated once to calculate Pass@1. Matches against Stockfish, using only one sampling iteration per match, evaluated the legality of our model's moves. Figure~\ref{fig:elototal} shows our model's results. Despite fluctuations from incorporating more endgame strategies, the model consistently achieves over 90\% move legality. The legality remains stable against opponents of varying strengths.
\noindent
\paragraph{Elo rating.}Table~\ref{tab:elo chessegpt} shows our model's performance in 100 rounds each against Stockfish at skill levels $0,1,2$etc., computing Elo ratings. With $temperature$ and $top_p$ parameters were both set at $0.7$, and $top_k$ was set at $50$. we used up to $10$ sampling iterations, performing the move upon obtaining a legal one. Our model achieves an Elo score of about $1788$, positioning it at the top of amateur chess performance.


\subsection {Eval Set Accuracy and Actual Games}
Figure~\ref {fig:elo-best} shows that within the evaluation set, an increase in Best Move accuracy correlates with Elo rating gains. A significant Elo rating jump occurs when the model's Legal Move accuracy reaches 99.8\%. This increase is due to the reduction in errors after the model learns to generate legal moves, reinforcing that continuous error correction and learning the correct moves significantly improve Elo ratings.

\subsection{Compare with Other LMs}

\paragraph{General Policy.} General Policy is proposed by ChessGPT~\cite{feng2023chessgpt}. Table~\ref{tab:gpcompare} showcases the results, delineating the effectiveness of various models in identifying the most fitting move for the black chess piece. 

\paragraph{Win Rating.}We conduct matches between ChessLLM and other Language Models (LMs) such as LLAMA~\cite{touvron2023llama}, RedPajama~\cite{together2023redpajama}, ChessGPT-Base~\cite{feng2023chessgpt}, and ChessGPT-Chat~\cite{feng2023chessgpt}, calculating their respective win rating. As other models cannot guarantee the legality of the moves they generate, we bring in Stockfish to aid in this process. Should the model fail to produce a valid move even after 50 sampling efforts, a mechanism is employed wherein there's a 50\% chance of favoring either the best move identified by Stockfish or a randomly picked move from the list of all possible legal moves. Similarly, as ChessGPT is unable to generate the best move for the next step, we generate all legal moves through Stockfish and utilize their proposed general policy for selection, picking the most optimal move as recognized by the model.

\subsection{Impact of Token Quantity and Quality}
We have investigated the impact of data quantity and quality on the generation of legal moves. Figure~\ref{fig:elototal} Left presents the $Pass@1$ indicators for two groups of data. It can be observed that the model performance significantly improves with the addition of more high-quality data, supplementing the data beyond the original distribution. Figure~\ref{fig:elototal} Right presents an augmentation in the number of tokens, it is observed that the model's Elo rating experiences an enhancement. Concurrently, the enrichment of the model with data not within the distribution can expedite the elevation of the model's Elo rating.
\section{Conclusion}

In this paper, we convert chess to a text game and introduce a large-scale Fen-Best Move pair dataset. With the dataset, we propose the Large language model ChessLLM that can play a complete chess game. Considering the limitation of the evaluation set in out-of-distribution data, we propose the need to evaluate model capabilities in actual games. ChessLLM finally achieves an Elo rating of 1788 through the SFT method. In subsequent work, we will discuss how to improve ChessLLM by improving the data quality.
\section{Limitations}

In this study, we explored the problem of LLM playing chess games and found that with high-quality synthetic data of complete games, LLM can have the extrapolation and combat capabilities of chess games. In the future, we will continue to explore this capability by improving the data quality, RLHF, and self-play + MCTS so that LLM can become better at chess games.
Our ultimate goal is to enable LLM to excel in various games through high-quality game data.

\section{Ethics Statement}
In this research, we adhere to strict ethical guidelines and principles. The study has been designed and implemented with respect for the rights, privacy, and well-being of all individuals involved. Our findings and conclusions are reported accurately and objectively, avoiding any misrepresentation or manipulation of data. The entire process and outcomes are free from intellectual property and ethical legal disputes. 

\section*{Acknowledgments}
This work is supported by the National Natural Science Foundation of China (NO. 62102151), the Open Research Fund of Key Laboratory of Advanced Theory and Application in Statistics and Data Science, Ministry of Education (KLATASDS2305), the Fundamental Research Funds for the Central Universities.

\bibliography{custom}

\end{document}